\documentclass[conference]{IEEEtran}
\usepackage{url}
\usepackage{cite}
\usepackage{amsmath,amssymb,amsfonts}
\usepackage{graphicx}
\usepackage{textcomp}
\usepackage[vlined,algoruled,commentsnumbered]{algorithm2e} 
\usepackage{csquotes}
\usepackage{todonotes}
\usepackage{multirow}
\usepackage{siunitx}
\usepackage{coverpage}

\let\oldnl\nl
\newcommand{\nonl}{\renewcommand{\nl}{\let\nl\oldnl}}

\newlength\lenKwIn
\newcommand\myKwIn[1]{%
  \settowidth\lenKwIn{\KwIn{}}%
  \setlength\hangindent{\lenKwIn}%
  \nonl\hspace*{\lenKwIn}#1\\}

\newlength\lenKwOut

\DeclareMathOperator*{\argmin}{argmin}

\begin{document}

\title{Context-Aware Route Planning for Automated Warehouses}

\author{\IEEEauthorblockN{%
Jakub Hv\v{e}zda\IEEEauthorrefmark{1}\IEEEauthorrefmark{2},%
Tom\'a\v{s} Rybeck\'y \IEEEauthorrefmark{2},%
Miroslav Kulich\IEEEauthorrefmark{1} and%
Libor P\v{r}eu\v{c}il\IEEEauthorrefmark{1}}

\IEEEauthorblockA{\IEEEauthorrefmark{1}
Czech Institute of Informatics, Robotics, and Cybernetics\\
Czech Technical University in Prague\\ 
Prague, Czech Republic}

\IEEEauthorblockA{\IEEEauthorrefmark{2}
Department of Cybernetics\\
Faculty of Electrical Engineering\\
Czech Technical University in Prague\\ 
Prague, Czech Republic}}

\mauthor{Jakub Hv\v{e}zda, Tom\'a\v{s} Rybeck\'y, Miroslav Kulich, and Libor P\v{r}eu\v{c}il}
\published{{\it  Proceedings of 21st International Conference on Intelligent Transportation Systems (ITSC). 2018 IEEE Intelligent Transportation
Systems Conference}, Maui, 2018-11-04/2018-11-07. IEEE Intelligent Transportation Systems Society, 2018. p. 2955-2960. ISSN 2153-0017.}
\DOI{10.1109/ITSC.2018.8569712}
\original{https://ieeexplore.ieee.org/document/8569712}

\coverpage
\twocolumn

\maketitle

\begin{abstract}
In order to ensure efficient flow of goods in an automated warehouse and to guarantee its continuous distribution to/from picking stations in an effective way, decisions about which goods will be delivered to which particular picking station by which robot and by which path and in which time have to be made based on the current state of the warehouse. 
This task involves solution of two suproblems: (1) task allocation in which an assignment of robots to goods they have to deliver at a particular time is found and (2) planning of collision-free trajectories for particular robots (given their actual and goal positions).

The trajectory planning problem is addressed in this paper taking into account specifics of automated warehouses. 
First, assignments of all robots are not known in advance, they are instead presented to the algorithm gradually one by one.
Moreover, we do not optimize a makespan, but a throughput -- the sum of individual robot plan costs.

We introduce a novel approach to this problem which is based on the context-aware route planning algorithm~\cite{terMors2010}.
The performed experimental results show that the proposed approach has a lower fail rate and produces results of higher quality than the original algorithm. 
This is redeemed by higher computational complexity which is nevertheless low enough for real-time planning.

\end{abstract}


\section{Introduction}
\label{sec:intro}

\begin{figure}[tb]
  \centering
  \includegraphics[width=\columnwidth]{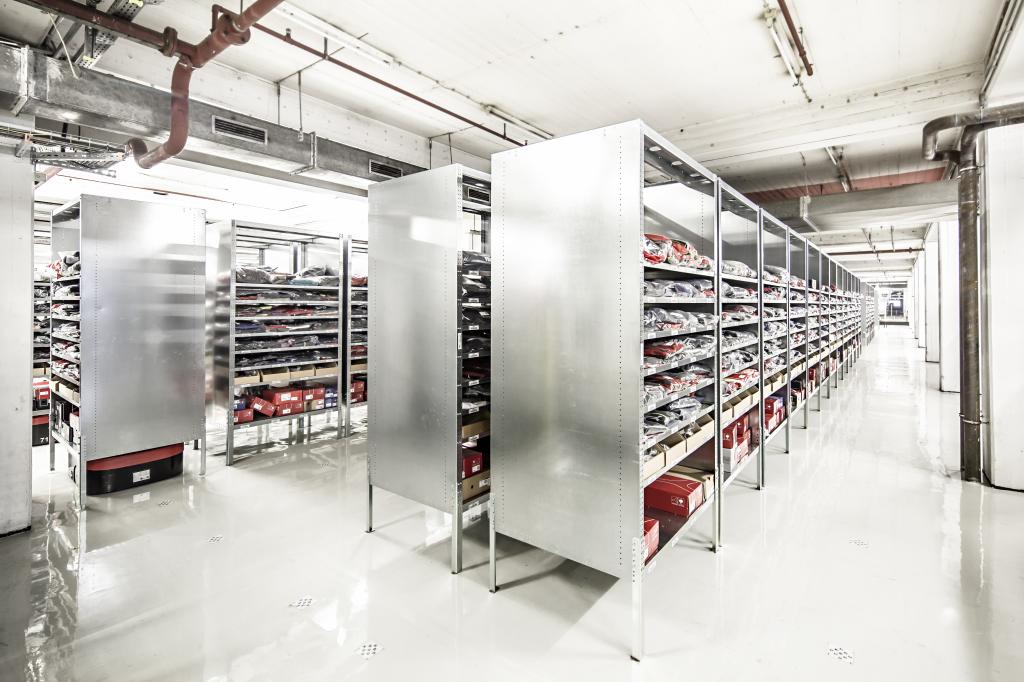}
    \caption{An automated warehouse: G-COM system by Grenzebach ({https://www.grenzebach.com}) in a costumer application.}
  \label{fig:gcom}
\end{figure}

The European market for e-commerce is growing rapidly, with more than 16\% just in the year 2014~\cite{europ_ecom_2014} and so is the same for the world market. 
With the growing markets, the need for larger warehouses and distribution centers and their automation increases~\cite{deloitte_2014}. 
In such facilities goods for the end-users or products in the business-to-business sector are stored, commissioned and shipped.
To manage the supply chains, many new warehouses have been erected, and more will follow (see Fig.~\ref{fig:gcom} for an example of such automated warehouse). 
Reliable figures are hard to come by, but for instance, a company such as DHL alone operates more than 2000 warehouses.
With the growing markets, the need for larger warehouses and their automation increases. 
Therefore the robotic and automation companies should be able to provide appropriate solutions, making scalable systems and scalable software mandatory.

One of the fundamental problems related to automated warehouses is trajectory planning and motion coordination for a fleet of robots distributing goods in a warehouse which has been theoretically studied from many perspectives since the 1980s, see~\cite{Parker2009} for a nice overview.
The problem (also formulated as the warehouseman's problem) has been proven to be PSPACE-complete~\cite{Hopcroft1984}.
The complexity of the problem can be reduced for the case where robots move on a predefined graph, nevertheless, it is still NP-hard~\cite{Goldreich11}, which means that optimal solutions cannot generally be found in a reasonable time for non-trivial instances (e.g., for a number of robots in order of tens).

Solutions to the problem consider either coupled or decoupled approaches. 
Coupled~(also called centralized) approaches consider a fleet of robots as a single multi-robot body with the number of degrees of freedom~(DoFs) equal to the sum of DoFs of individual robots.
Classical single-robot planning approaches can then be applied to plan motion of this multi-robot body in composite configuration space.
Because these algorithms are typically based on classical complete and optimal algorithms, their main advantage is that they warrant to find a solution if it exists and report its non-existence if it does not exist~\cite{Latombe1991},~\cite{Lavalle1998},~\cite{Ryan2008}.
This, however, leads to the main disadvantage of these algorithms which is the exponential computational time in the dimension of the composite configuration space.
The appropriate use cases for this type of algorithms are thus problems that consist of a small number of robots. 

On the other hand, algorithms that use the decoupled approach typically provide solutions much faster than coupled planners for the price of the solutions being sub-optimal.
In addition to this, decoupled algorithms generally do not guarantee the completeness and may suffer from deadlocks.
These approaches can be typically split into two categories - path coordination and prioritized planning.
Path coordination tunes velocities of individual robots along precomputed trajectories to avoid collisions of these robots~\cite{LaValle1998_OMP},~\cite{Simeon2002}.
Prioritized planning assigns each agent a priority according to which an ordering is made for the agents.
The planning algorithm then plans trajectories for individual robots according to this ordering, while considering already planned robots as moving obstacles that have to be avoided.~\cite{VandenBerg2005},~\cite{Bennewitz2001},~\cite{Cap2015}. 

\cite{terMors2010} presents a similar idea, but codes information about trajectories of robots with higher priorities into the planning graph rather than into the planning algorithm itself. 
It does so by constructing a resource graph, where each resource can be for example a node of the original graph or intersection graph edges.
Every such resource holds information about time intervals in which it is not occupied by already planned robots.
An adaptation of the A* algorithm is used on this graph to find the shortest path through these intervals (called free time windows) to obtain a path that avoids all already planned robots.

\cite{Gawrilow2008} deals with a real-life problem of routing vehicles in Container Terminal Altenwerder in Hamburg harbor.
They used a similar approach to keeping a set of free time windows for path arcs in the graph.
Their algorithm contains a preprocessing of the graph for the use of specific vehicles followed by computation of paths for individual vehicles on this preprocessed graph.

Another similar approach is presented in \cite{spatioTempA*}, where each robot looks for a viable path in a 2D spatial grid map and checks for collisions with moving obstacles using a temporal occupancy table.
\cite{Zhang2015} adopted a similar approach to \cite{terMors2010} with enhanced taxiway modeling approach to improve performance on airport graph structures.
\cite{terMorsComparison} compares Context-Aware Route Planning~(CARP)~\cite{terMors2010} with a fixed-path scheduling algorithm using $k$ shortest paths~\cite{Yen1971} and a fixed-path scheduling algorithm using $k$ disjoint paths~\cite{Suurballe1974}.
The experiments show that the CARP algorithm is superior in all measured qualities.

\cite{terMorsHeuristics} compares several heuristic approaches of assigning priority to robots and concludes that the heuristics which plans longest paths first perform best when a makespan is to be minimized. 
A greedy best-first heuristics provides best results regarding joint plan cost. However, its downside is that it calls the planning algorithm for all yet unplanned robots in every round and it is thus very time-consuming.

In this paper, we consider the planning problem from the perspective of usability in automated warehouses (AWs).
The key difference in AWs is that the assignments are not known all at once at the time of planning, but they instead come in time sequentially as requests for goods to be delivered are being received.
This emphasizes a quick time to compute a plan for newly requested robots while minimizing the sum of time it takes for all robots in the system to complete their plans.
We present a novel approach that can calculate paths for newly added robots to the system while maximizing its throughput.

The rest of the paper is organized as follows. 
The multi-robot path-finding problem for automated warehouses is presented as well as the used terms are defined in Section~\ref{sec:problem}.
The proposed planning algorithm is described in Section~\ref{sec:alg}, while performed experiments, their evaluation, and discussion are presented in Section~\ref{sec:exp}.
Finally, Section~\ref{sec:conclusion} is dedicated to concluding remarks and future work.

\section{The problem}
\label{sec:problem}
Assume a finite weighted connected graph $G(V,E)$ and a set ${\cal A} = \left(a_1,a_2,...,a_{k-1}\right)$ of robots each of which has already planned a trajectory in $G$ without collisions. 
The aim is to find a non-colliding trajectory for a robot $a_k$ from its start position to the given target position while optimizing a global cost function, possibly without replanning all robots.


We employ the CARP algorithm~\cite{terMors2010} for the planning of individual robots, so we use its assumptions on the graph structure as follows.
The infrastructure the robots plan on is modeled as a resource graph $G_R = (R,E_R)$ where resources $R$ can be the original nodes, edges, intersections of edges, etc.
The path the robot can follow is restricted to edges $E_R$ which limit transitions of robots such that a robot can move from a resource $r_1$ to a resource $r_2$ only if $r_2$ is a neighbour of $r_1$ in the graph $G_R$, i.e. the edge $(r_1,r_2) \in E_R$.
Each resource also has two main attributes associated with it.
These are capacity $c(r)$ which corresponds to the maximum number of robots that can occupy the resource $r$ at the same time and duration $d(r)>0$ that represents the minimal time it takes the robot to traverse the given resource $r$.
Plans for every robot then contain not only a sequence of resources on its path but also time intervals during which the robot visits them.
The idea is that if the infrastructure is modeled this way, collision avoidance is shifted into deciding which resources at what time the robots can visit, rather than finding paths without collision in space and time, which is PSPACE-complete~\cite{Hopcroft1984}.

\section{Algorithm}
\label{sec:alg}

\newcommand{\IO}{{\cal A}}
\newcommand{\Neib}{{\cal N}}
\newcommand{\Res}{{\cal T}^{new}}

As mentioned, the proposed algorithm employs CARP~\cite{terMors2010}.
It was chosen because of its ability to plan sequentially one robot at a time while keeping the information about movement of other robots in a compact and easy to update manner. 
More specifically, CARP uses the infrastructure described above to find paths in a resource graph, where each node of the infrastructure keeps information about its capacity and occupancy in given time windows.
This allows it to use a modified A* algorithm that finds a free time window in the start resource corresponding to a start node and attempts to find a path through these time windows to a free time window in the resource that corresponds to the goal node.

The proposed algorithm aims to generate a trajectory for an robot $a_k$ assuming that trajectories for $k-1$ robots are already planned which can possibly lead to modification of those planned trajectories.
The main idea is to iteratively build a set of robots whose trajectories mostly influence an optimal trajectory of $a_k$, see Algorithm~\ref{alg:alg}.
The algorithm maintains two structures:  
\begin{itemize}
\item $\Neib$ -- a set of robots in the neighborhood of $a_k$ which is initially set to contain $a_k$ (line~\ref{s4}), and
\item $\IO$ -- a sequence of robots not in $\Neib$. This sequence initially stores the order in which trajectories of the robots $a_1 \dots a_{k-1}$ were generated (line~\ref{s1}).   
\end{itemize}
Moreover, cost of the best solution that has been found so far is set to a high number.

New robots are iteratively added to the neighborhood in the loop starting at line~\ref{s6}.
The robot $a_b \in \IO$ which minimizes the distance to the neighborhood is found at each iteration first, where the distance of a robot to a set is defined as the distance of the robot to the closest robot in the set. 
The distance between two robots is then determined as the average Euclidean distance of robots' positions at discrete time steps:
$$ d(a_i, a_j) = \frac{\sum_{\tau=\tau_{S}}^{\tau_{G}}{\Big| t_i(\tau), t_j(\tau) \Big| }}{\tau_G-\tau_S},$$
where $t_i(\tau)$ is a position of $a_i$ at time $\tau$ if it follows the trajectory $t_i$,  
$t_j(\tau)$ is a position of $a_j$ at time $\tau$ if it follows the trajectory $t_j$,
and $\left<\tau_{S}, \tau_{G} \right>$ is a time interval when $a_i$ or $a_j$ moves.
Note that trajectories of robots in $\IO$ are initially taken from the input set $\cal T$ while an initial trajectory of $a_k$ is determined as the shortest path between its start and goal positions on $G$ making use of the A* algorithm (line \ref{s3}).
These initial trajectories are updated as soon as new plans are found at the next steps of the algorithm. 

\begin{figure}[htb]
    \centering
    \includegraphics[width=0.8\columnwidth]{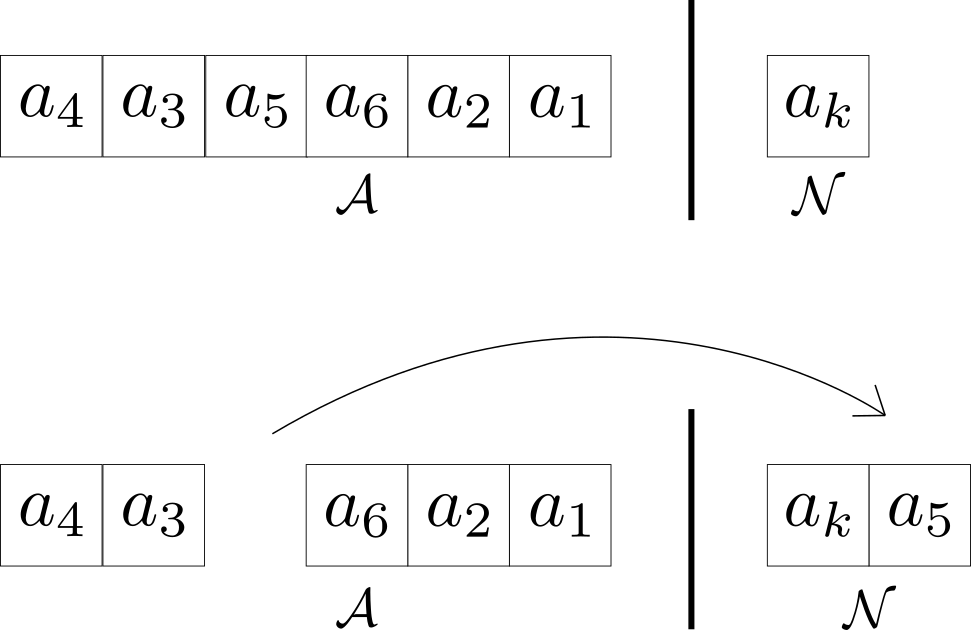}
    \caption{Adding an robot into a set of neighbors.}
    \label{fig:add_agent}
\end{figure}

The found closest robot $a_b$ is then removed from the sequence $\IO$ (line~\ref{s8}), added to a set of neighbors $\Neib$ (line~\ref{s8}) and new trajectories for $\IO$ are computed by the CARP algorithm from the scratch (line~\ref{s10}) as demonstrated in Fig.~\ref{fig:add_agent}.

All possible permutations of robots in $\Neib$ are considered next (line~\ref{s11}). 
A set of trajectories $\cal P$ is determined by the CARP algorithm taking into account trajectories $\cal C$ for each such permutation (line~\ref{s12}). 
This is realized by running CARP for $\Neib$ on a resource graph with free windows generated by CARP when computing $\cal C$ at line ~\ref{s10}. 
The set $\cal P$ is then added to $\cal C$, and the cost of this solution is computed (line~\ref{s13}) and compared with the best solution found till now (line~\ref{s14}).
If the new solution is better than the currently best, it is stored together with its cost (lines~\ref{s15} and \ref{s16}).
The best-found solution is finally reported at line~\ref{s17}.


\SetKwFor{Times}{}{times do}{end}
\LinesNumbered
\DontPrintSemicolon

\begin{algorithm}[ht]
\caption{Plan update}
\label{alg:alg}
\KwIn{$G=(V,E)$ -- a graph}
\myKwIn{${\cal T} = \left\{ t_i\right\}_{i=1}^{k-1}$ -- already planned trajectories}
\myKwIn{$V_S$ -- start position}
\myKwIn{$V_G$ -- goal position}
\myKwIn{$M$ -- size of a neighborhood}
\KwOut{${\cal T}^{new} = \{ t_i\}_{i=1}^k $ -- updated set of trajectories}
\nonl\hrulefill\\
$\IO \leftarrow \left<a_1, a_{k-1}\right>$\label{s1}\\
$t_k\leftarrow$ shortest path from $V_S$ to $V_G$ in $G$\label{s3}\\
$\Neib \leftarrow \{ a_k \}$ \label{s4}\\
$best \leftarrow \infty$\label{s5}\\
\Times{$M-1$} { \label{s6}
    $a_b = \argmin_{a_i\in \IO } d(a_i, \Neib)$\label{s7}\\
    $\IO \leftarrow \IO \setminus \{a_b\}$ \label{s8}\\
    $\Neib \leftarrow \Neib \cup \{a_b\}$\label{s9}\\
    ${\cal C} \leftarrow CARP(G, \IO)$\label{s10}\\
    \ForEach{$\pi \in \Pi(\Neib)$}{ \label{s11}
        ${\cal P} \leftarrow CARP(G, \pi, C)$\label{s12}\\
        $ c \leftarrow cost({\cal C} \cup \cal P)$ \label{s13}\\
        \If{$c<best$} { \label{s14}
            $best \leftarrow c$ \label{s15}\\
            $\Res \leftarrow {\cal C} \cup {\cal P}$ \label{s16}\\
        }        
    }
    \Return $\Res$ \label{s17} \\
}

\end{algorithm}

Calculation of computational complexity of the proposed algorithm is based on the fact that CARP for $n$ robots comprises $n$ calls of A*. 
We call CARP $M-1$ times at line~\ref{s10} gradually for $k-2, k-3, \dots k-M-1$ robots which leads to $\frac{M}{2}(2k-M-3)$ calls of A*.
Similarly, CARP at line~\ref{s12} is called $\sum_{N=2}^{M}N!$ times which leads to $\sum_{N=2}^{M}NN!$ calls of A*.

The total number of A* calls can be significantly reduced in two ways.
Firstly, when calling CARP at line~\ref{s10} after removal of $a_b$ not all trajectories have to be recomputed. 
We can instead preserve trajectories of robots which were in $\IO$ before $a_b$ as they are not influenced by $a_b$. 
Only trajectories of robots behind $a_b$ in $\IO$ have to be recomputed which leads to a reduction of A* calls by 50\% in average.

\begin{figure}[htb]
    \centering
    \includegraphics[width=0.4\columnwidth]{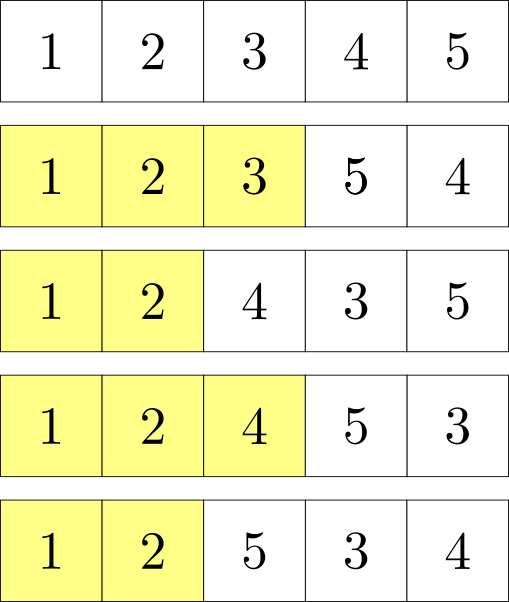}
    \caption{First permutations of five elements in the lexicographic order. The yellow plans can preserved.}
    \label{fig:perms}
\end{figure}

The second reduction is similar.
If the permutations are generated in a lexicographic order at line~\ref{s11} then two consecutive permutations have typically a big joint head as depicted in Fig.~\ref{fig:perms}.
Plans of robots in that head can be preserved while recomputation has to be done only for the rest. 
Assuming neighborhood size $|{\cal N}|=4$, instead of calling A* $4\times 4! = 96$ times, only 64 calls is performed which is $67\%$.    
The reduction is even greater for $|{\cal N}|=5$: 325 calls instead of 600 which is $54\%$, while only $45\%$ (1956 instead of 4320) A* calls are needed for $|{\cal N}|=5$.


\section{Experiments}
\label{sec:exp}

The experiments were performed on a computer equipped with Intel Xeon E5-2690.
The maps the experiments were performed on were created to show how the proposed algorithm performs depending on the density of the given graph, specifically the number of edges.
The set of 21 maps was generated by creating a minimum spanning tree of a $20\times 20$ grid map and then iteratively adding a given number of original edges to it until the original grid was recreated.
Furthermore, 500 different random assignments were generated for a fleet of 100 robots by randomly sampling start and goal nodes for each robot.
Each of these assignments was tested on all 21 maps.

The goal of the experiments was to test how the proposed algorithm scales with the number of edges in the graph as robots are sequentially added to the system.
For comparison we ran the original CARP algorithm with no change to priorities of the robots, i.e., priorities were set randomly.
Because this approach proved to have a high failure rate, we introduced two variants that after each planning attempt randomly shuffled the robot order 10 and 100 times ({\tt CARP10}, {\tt CARP100} respectively) and tried planning again from scratch.
The best plan regarding the sum of the number actions of individual robots was considered as the result.
We also added CARP with the longest first heuristic to determine robot priorities as introduced in~\cite{terMorsHeuristics} to the comparison as {\tt LF}.
The proposed algorithm was run in several variants that differ in the parameter $\cal M$ specifying the size of the neighborhood.
For the parameters 4,5,6 ({\tt Proposed\_4}, {\tt Proposed\_5}, {\tt Proposed\_6} respectively) all the permutations of the neighborhood were considered.
For the parameter 10 ({\tt Proposed\_10}) a 150 different neighborhood permutations were chosen randomly to decrease computational complexity.

\begin{figure}[htb]
  \includegraphics[width=\columnwidth]{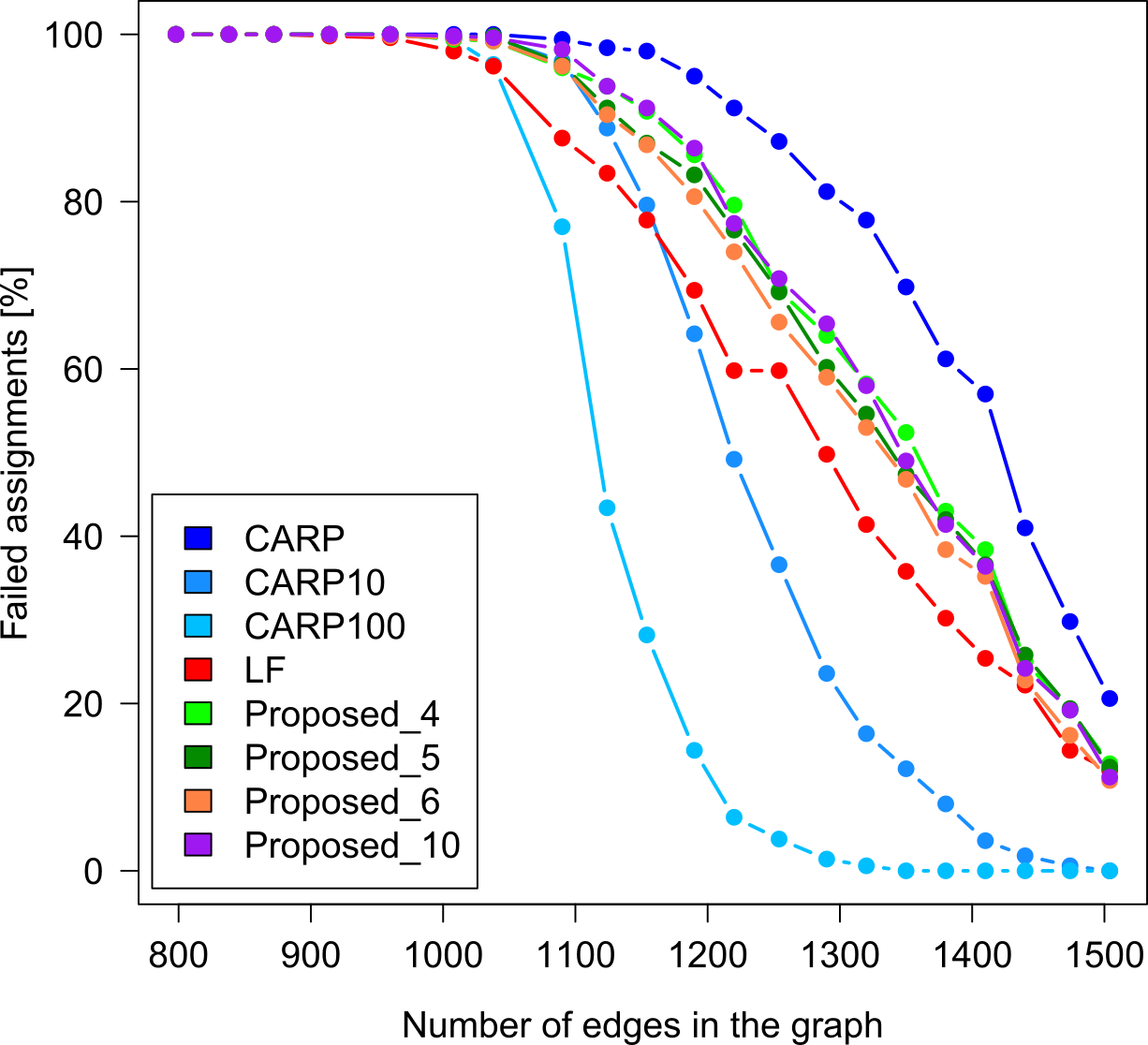}
  \caption{Algorithms fail rate}
  \label{fig:fail_rate}
  \end{figure}

\begin{figure}[htb]
\includegraphics[width=\columnwidth]{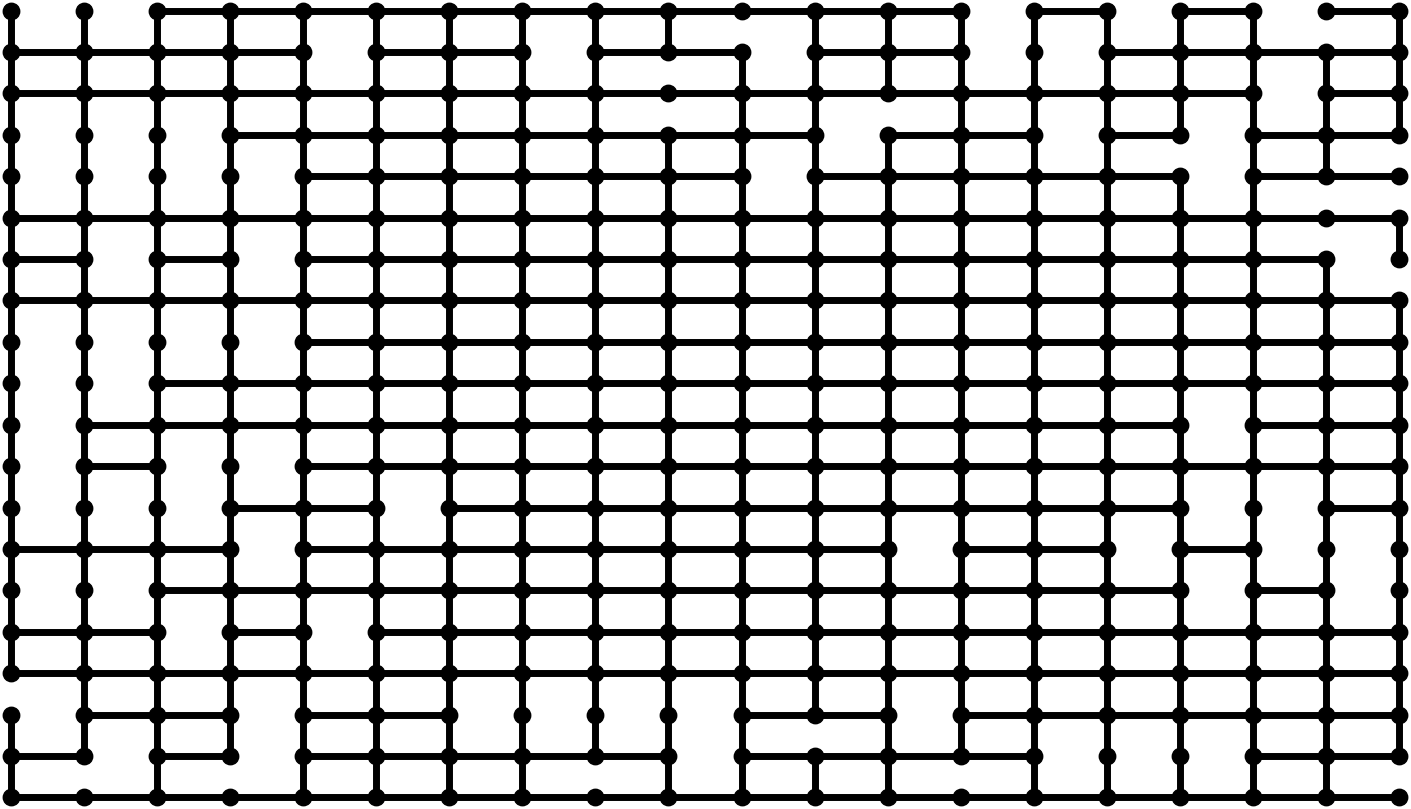}
\caption{Example of an experimental map}
\label{fig:map_15}
\end{figure}

\begin{figure}[htb]
\includegraphics[width=\columnwidth]{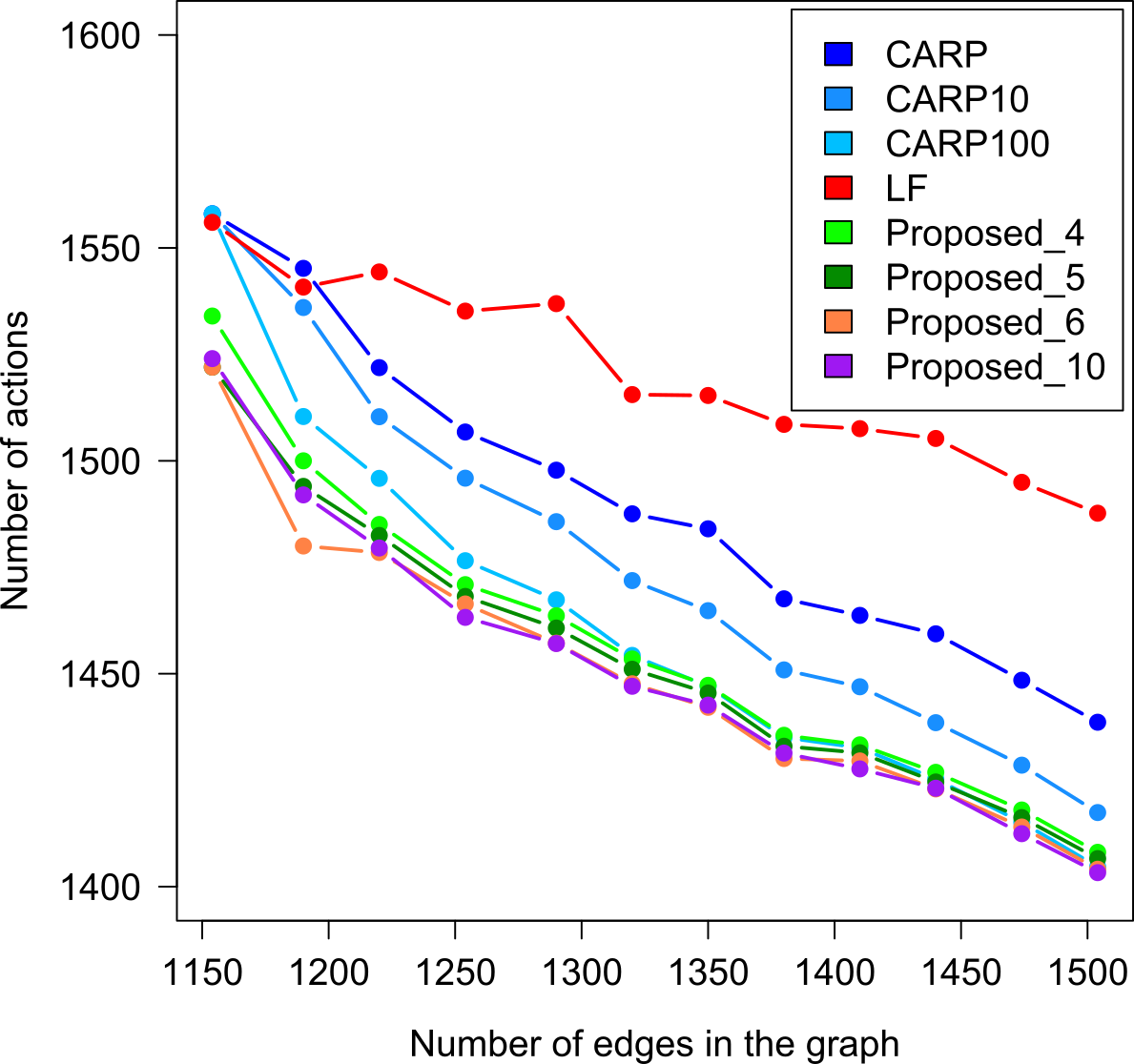} 
\caption{Sum of actions performed by all robots}
\label{fig:length}
\end{figure}

Orders for all 100 robots were presented sequentially to all algorithms on all maps and assignments in the first experiment.
Failure rate (Fig.~\ref{fig:fail_rate}) and the number of actions of all robots were observed (Fig.~\ref{fig:length}).
In the second experiment, the algorithms were sequentially given one robot at a time at each iteration for all 500 assignments on a map shown in Fig.~\ref{fig:map_15} with the goal of showing how much time it takes to generate a plan for each algorithm after adding robot into the system.

The results for the failure rate of the algorithms can be seen in Figure \ref{fig:fail_rate}.
It can be seen that the proposed algorithm has a fail rate in between the fail rates of {\tt CARP} and {\tt CARP10}.

The Figure \ref{fig:length} shows the average overall quality of the plan from the first experiment measured as the total number of actions of all robots.
The graph shows that all versions of the proposed algorithm perform similarly to each other with {\tt Proposed\_10} having the best results.
Compared to the basic CARP algorithm the proposed algorithm has up to 65 less number of total actions performed across all robots in {\tt Proposed\_6} variant (on the 10th map) and up to 35 fewer actions for {\tt Proposed\_6} and {\tt Proposed\_10} on the original grid map.
Moreover, all variants of the proposed algorithm outperform {\tt CARP10} and are at least comparable to {\tt CARP100} which is much more time-consuming.
It can also be noticed that {\tt LF} generates worst results.  
It is not much surprising as it was designed to optimize a makespan.

\begin{figure}[htb]
\includegraphics[width=\columnwidth]{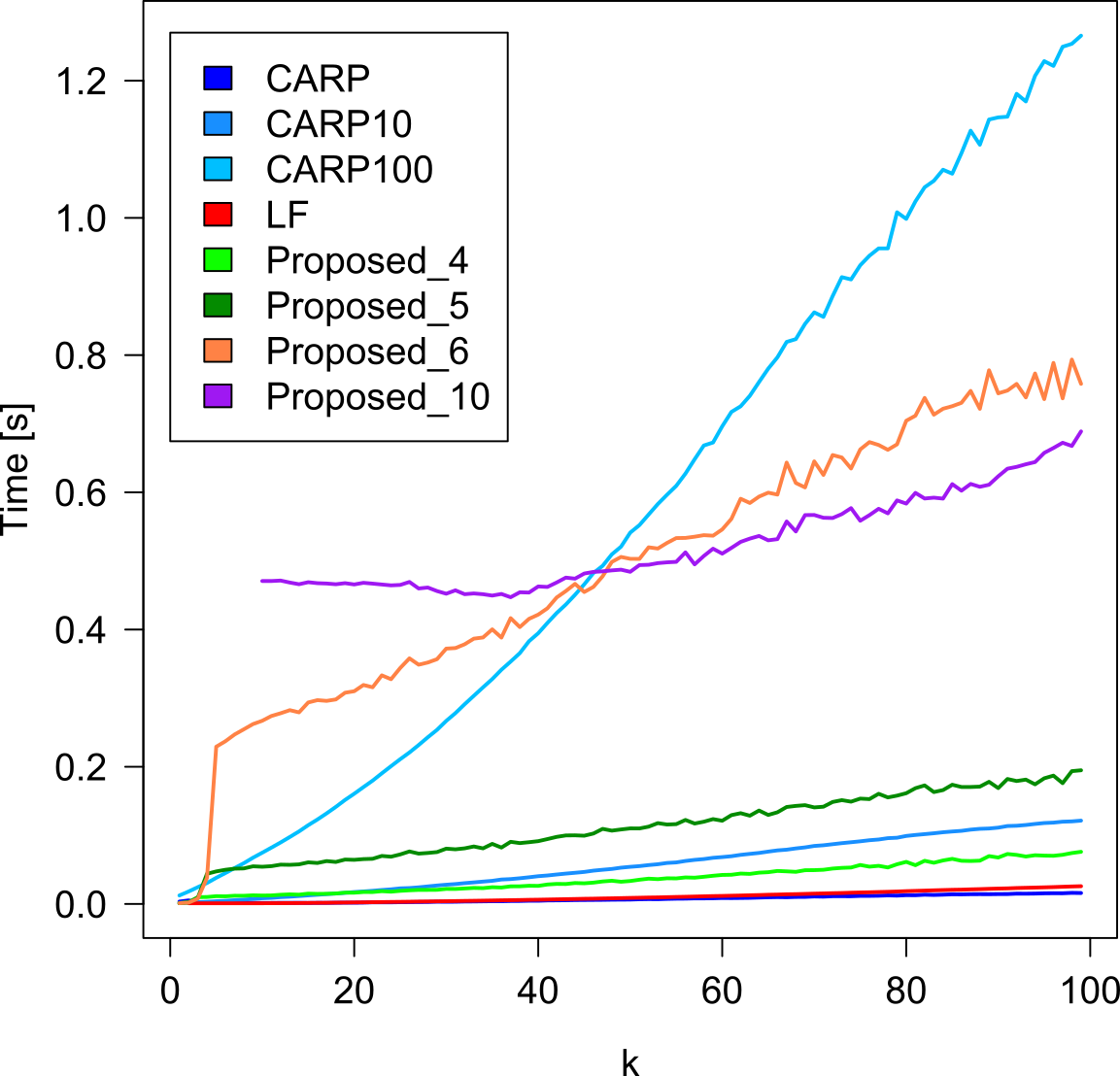} 
\caption{Time to plan $k^{th}$ robot}
\label{fig:plan_one}
\end{figure}

The results of the second experiment are presented in Fig.~\ref{fig:plan_one} which shows the required time to find a trajectory for $k$-th robot considering trajectories of robots $1\dots k-1$ are already planned. It is evident that {\tt CARP100} is the slowest of all tested algorithms with {\tt Proposed\_6} as the second slowest.
It is worth noticing that for {\tt Proposed\_6} the time of testing all permutations of the neighborhood took longer than replanning of the rest of the robots.
The table \ref{table:time_one} shows the actual times to plan 50th and 100th robot in this experiment.

From the results as a whole we can see that {\tt Proposed\_4} and {\tt Proposed\_5} show the best ratio between the quality of the found plan and the time required to compute it.
Additionally, if the time requirements for planning are not as tight, it is possible to run a more demanding version of the algorithm to increase the quality of the solution.
To increase the overall success rate of the algorithm, it is possible to combine the proposed algorithm with a version that has a higher success rate, such as {\tt CARP100} or even higher in case the proposed algorithm does not find a solution.

\bgroup
\def\arraystretch{1.6}
\begin{table}[]
\centering
\caption{Times of addition for $50^{th}$ and $100^{th}$ robot}
\label{table:time_one}
\begin{tabular}{|p{10em}|S|S|}
\hline
\multirow{2}{*}{Algorithm} & \multicolumn{2}{c|}{Time for $k^{th}$ robot {[}ms{]}} \\ \cline{2-3} 
                           & {$50^{th}$ robot}             & {$100^{th}$ robot}             \\ \hline
Proposed\_4                & 3.23                   & 7.6                     \\ \hline
Proposed\_5                & 10.86                  & 19.48                   \\ \hline
Proposed\_6                & 50.58                  & 75.81                   \\ \hline
Proposed\_10               & 43.84                  & 54.10                   \\ \hline
CARP                       & 0.63                   & 1.60                    \\ \hline
CARP10                     & 5.29                   & 12.15                   \\ \hline
CARP100                    & 52.08                  & 126.57                  \\ \hline
LF                         & 0.85                   & 2.59                    \\ \hline
\end{tabular}
\end{table}
\egroup

\section{Conclusion}
\label{sec:conclusion}

In this paper, we present a novel algorithm for planning in automated warehouses that consider specific requirements in these environments for sequential addition of robots into the system while optimizing the total number of actions all robots must perform.
The proposed algorithm employs the standard CARP algorithm for the planning of individual robots, which is one of the best practical algorithms nowadays.
The experimental results show that the proposed approach finds better solutions than the original CARP algorithm after several random shuffles of the robots' priorities while requiring significantly less computational time for adding individual robots into the system.
Moreover, it is much faster than {\tt CARP100} which produces similar results.

The only drawback of the algorithm is the success rate of finding the solution, and thus the future work should focus on its improvement.
One approach could be the use of deep learning to better determine either robot priorities or the set of robots mostly influencing the planned robots which need to be replanned.
Another route of improvement would be to adapt the system for the use of heterogeneous robots and more importantly humans, who need to have the highest priority to have their trajectories planned to minimize the time humans spend in the warehouse to eliminate the risk of injury.

\section*{{Acknowledgement}}

\noindent This work has been supported by the European Union's Horizon 2020 research and innovation programme under grant agreement No 688117, by the Technology Agency of the Czech Republic under the project no.~TE01020197 \enquote{Centre for Applied Cybernetics}, and
by the European Regional Development Fund under the project Robotics for Industry 4.0 (reg. no. CZ.02.1.01/0.0/0.0/15 003/0000470). 
The work of Jakub Hv\v{e}zda was also supported by the Grant Agency of the Czech Technical University in Prague, grant No.~SGS18/206/OHK3/3T/37.

\bibliographystyle{IEEEtran}
\bibliography{IEEEabrv,./main}

\end{document}